\documentclass[11pt,a4paper]{article}
\usepackage[hyperref]{emnlp2020}
\usepackage{times}
\usepackage{latexsym}

\usepackage{url}

\usepackage{microtype}

\usepackage{epsfig}
\usepackage{graphicx}
\usepackage{xcolor}

\usepackage{footnote}
\makesavenoteenv{tabular}
\makesavenoteenv{table}
\makesavenoteenv{minipage}
\makesavenoteenv{figure}

\usepackage{booktabs}
\usepackage{makecell}
\usepackage{color}
\usepackage{amsmath}
\usepackage{amssymb}
\usepackage{multirow, varwidth}
\usepackage{dblfloatfix}
\usepackage{verbatim}
\makeatletter
\newif\if@restonecol
\makeatother

\usepackage[ruled,vlined]{algorithm2e}
\usepackage{xr}
\usepackage[amsmath,thmmarks]{ntheorem}

\DeclareRobustCommand\onedot{\futurelet\@let@token\@onedot}
\def\onedot{. }
\def\eg{\emph{e.g}\onedot} 
\def\ie{\emph{i.e}\onedot}

\definecolor{myred}{RGB}{255, 86, 93}
\definecolor{myblue}{RGB}{86, 125, 255}

\aclfinalcopy %
\def\aclpaperid{419} %

\newcommand{\norm}[1]{\left\lVert#1\right\rVert}

\title{\textit{Will I Sound Like Me}? \\ Improving Persona Consistency in Dialogues \\ through Pragmatic Self-Consciousness}

\author{
    Hyunwoo Kim \qquad Byeongchang Kim \qquad Gunhee Kim \\
    Department of Computer Science and Engineering\\
    Seoul National University, Seoul, Korea \\
    {\tt \small{ \{hyunw.kim, byeongchang.kim\}@vl.snu.ac.kr gunhee@snu.ac.kr }} \\
    \url{https://vl.snu.ac.kr/projects/consistency}
}

\date{}

\begin{document}
\maketitle

\newcommand{\hmm}[1]{{\color{magenta}{\small\bf\sf [Hmm..: #1]}}}
\newcommand{\new}[1]{{\color{teal}{\small\bf\sf [NEW: #1]}}}
\newcommand{\bc}[1]{{\color{blue}{\small\bf\sf [@Byeongchang: #1]}}}

\begin{abstract}

    We explore the task of improving persona consistency of dialogue agents. %
    Recent models tackling consistency often train with additional Natural Language Inference (NLI) labels or attach trained extra modules to the generative agent for maintaining consistency.
    However, such additional labels and training can be demanding.
    Also, we find even the best-performing persona-based agents are insensitive to contradictory words.
    Inspired by social cognition and pragmatics, we endow existing dialogue agents with \textit{public self-consciousness} on the fly through an imaginary listener.
    Our approach, based on the Rational Speech Acts framework \cite{Frank:2012:Science}, can enforce dialogue agents to refrain from uttering contradiction. %
    We further extend the framework by learning the distractor selection, which has been usually done manually or randomly.
    Results on Dialogue NLI \cite{Welleck:2019:ACL} and PersonaChat \cite{Zhang:2018:ACL} dataset show that our approach reduces contradiction and improves consistency of existing dialogue models.
    Moreover, we show that it can be generalized to improve context-consistency beyond persona in dialogues.
\end{abstract}

\section{Introduction}
\label{sec:intro}

In the study of dialogue agents, \textit{consistency} has been a long-standing issue.
To resolve this, much research has been conducted to endow dialogue agents with \textit{personas}.
\citet{Li:2016:ACL} propose to encode persona in embeddings
and \citet{Zhang:2018:ACL} introduce a persona-conditioned dialogue dataset.
On top of these works, many efforts have been made to improve consistency.

\begin{figure}[t] \begin{center}
    \vspace{-5pt}
    \includegraphics[width=\linewidth]{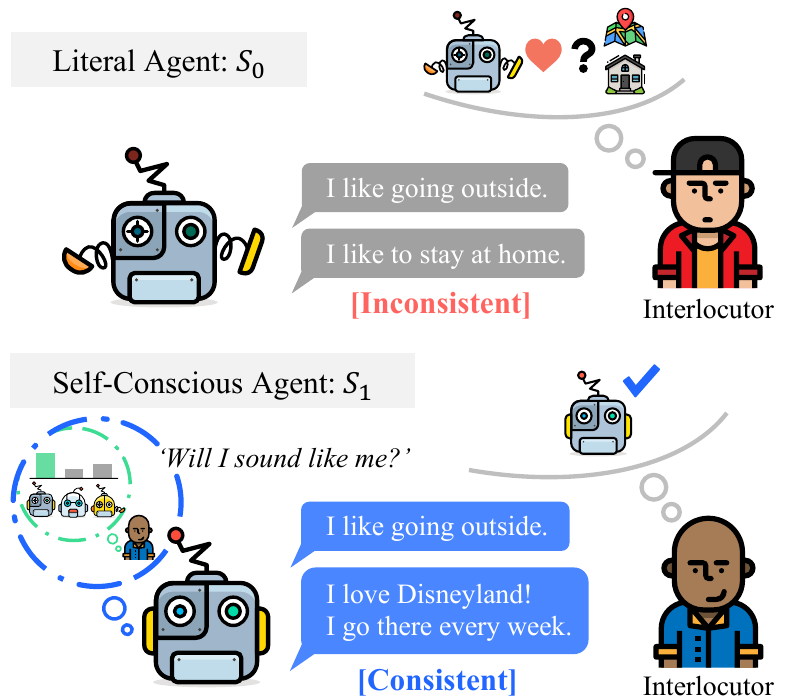}
    \caption{Illustration of the consistency issue in dialogue. While a literal dialogue agent ($S_0$) fails to deliver a consistent persona, our self-conscious agent ($S_1$) does so,
             by modeling an imaginary listener.
    Icons are designed by Nhor Phai and Vincent Le Moign.}
    \label{fig:front}
    \vspace{-15pt}
\end{center} \end{figure}

In spite of such recent significant progress, there is much room for improving persona-based dialogue agents.
We observe that even the best performing persona-based generative models \cite{See:2019:NAACL, Wolf:2019:arXiv, Roller:2020:blender}
are highly insensitive to contradictory words, and thus fail to deliver consistent persona to the interlocutor (Figure \ref{fig:front}).
Also, extra modules other than the generative model is often required for improving consistency.
Recent works on consistency in persona-based dialogue actively adopt the NLI-based approach \cite{Welleck:2019:ACL, Song:2019:arXiv, Li:2020:ACL, Song:2020:ACL}, which have the following prerequisites.
First, they require labeled pairs of persona sentences and dialogue utterances with three categories: entailment, neutral, and contradiction.
Next, methods with NLI models for rating the agent's consistency also need to train them separately with those labels.

In this work, we step back from this NLI-based supervised approach and ponder: \textit{how do humans maintain consistency?}
We humans never learn how to be consistent.
Instead, we have an innate drive for consistency to hold our beliefs and behavior in harmony \cite{Festinger:1962:dissonance}.
If so, how do we know we are consistent or not?
We do not ask others. We ask ourselves by predicting how we are perceived by others.
\textit{Public self-consciousness} is this awareness of the self as a social object that can be observed and evaluated by others \cite{Fenigstein:1975:JCCP}. %
We particularly emphasize that public self-consciousness is not equivalent to the philosophical self-consciousness (or self-awareness)\footnote{\url{https://plato.stanford.edu/entries/self-consciousness/}}.
Simply put, public self-consciousness is the concern about how oneself will be perceived by others, as opposed to the philosophical state of being conscious of self-existence.

According to \citet{Doherty:1991:JPers}, people with high public self-consciousness tend to act more consistent with known information about themselves.
They care deeply about how others will evaluate them and have a strong tendency to avoid negative evaluations \cite{Fenigstein:1975:JCCP}.
Since inconsistency is condemned by others, one who has high public self-consciousness will try more to maintain consistency.
In order to predict how we are perceived,
we rely on abstract models of others \cite{Gopnik:1992:Mind}
and simulate others' reactions based on imagination \cite{Hassabis:2013:Cerebral}.
Inspired by this, our intuition is that self-consciousness through an imaginary listener will let dialogue agents better maintain consistency.

Modeling a listener has been one of the main topics in computational pragmatics.
Our work extends this long line of work in cognitive science by making use of the Bayesian Rational Speech Acts framework \cite{Frank:2012:Science},
which has been originally applied to improving informativeness of referring expressions.
Since personas ought to express who we are,
we adopt this framework for dialogue agents by regarding personas as targets that should be conveyed to the interlocutor.
As the agent tries to generate tokens that help the imaginary listener identify the agent's persona, it can lastly generate more consistent utterances.
In summary, we take inspiration from social cognition and pragmatics to endow generative agents with self-consciousness,
which makes them imagine the listener's reaction and incorporate it to the generation process for improving consistency.
Our major contributions can be outlined as follows:

(1) %
We propose an orthogonally applicable approach for any persona-based generative agents to improve consistency without the use of additional consistency labels and training.
Moreover, it is even generalizable to improve context-consistency beyond persona in dialogue.
(2) We extend the Rational Speech Acts framework \cite{Frank:2012:Science} with two new technical features:
(i) a learning method for distractor selection (\eg other samples different from the given target \cite{Andreas:2016:EMNLP}), which has been usually done manually or randomly,
and (ii) a different update for the listener's world prior that better preserves information of previous states.

(3) Our approach improves consistency of three recent generative agents \cite{See:2019:NAACL, Wolf:2019:arXiv, Roller:2020:blender} over Dialogue NLI \cite{Welleck:2019:ACL} and PersonaChat \cite{Zhang:2018:ACL}.
Along with large reduction in contradiction, the utterance accuracy significantly increases too.

\section{Related Work}
\label{sec:related}

\textbf{Persona \& Consistency in Dialogue}.
\citet{Li:2016:ACL} learn personas in embeddings. %
\citet{Zhang:2018:ACL} release the \textit{PersonaChat} dataset, a chitchat dialogue set involving two interlocutors each playing their given persona.
\citet{Madotto:2019:ACL} use meta-learning to adapt to new personas with few dialogue samples.
\citet{Liu:2020:ACL} use reinforcement learning to enhance mutual persona perception.

Recent works use extra modules or NLI labels to improve consistency.
\citet{Shum:2019:arXiv} fill generated templates, and rank with a language model.
\citet{Zhang:2019:arXiv} use self-supervised feature extractors for generation.
\citet{Welleck:2019:ACL} annotate NLI labels to the PersonaChat dataset.
They train an NLI model and run pairwise comparison between candidates and persona to compute contradiction scores. %
The NLI approach is applied for coherence evaluation \cite{Dziri:2019:NAACL}, rewards to reinforcement learning agents \cite{Song:2019:arXiv},
finding inconsistent words \cite{Song:2020:ACL}, and unlikelihood training \cite{Li:2020:ACL}.
They require NLI labels on the target dialogue dataset; otherwise, sharp decrease in performance is observed, due to mismatch of data distribution \cite{Welleck:2019:ACL}.
Such dataset-specific NLI annotations and training NLI models can be costly and time-consuming.

Compared to previous methods,
the novelty of our approach is to improve consistency without NLI labels and extra modules.

\textbf{Pragmatics}.
Our approach belongs to the general family of Bayesian Rational Speech Acts (RSA) frameworks \cite{Frank:2012:Science} in pragmatics.
It has improved informativeness in a number of NLP tasks, including
reference games \cite{Andreas:2016:EMNLP}, image captioning \cite{Mao:2016:CVPR, Vedantam:2017:CVPR, Cohn:2018:NAACL},
instruction following \cite{Fried:2017:NAACL}, navigating \cite{Fried:2018:NeurIPS}, translation \cite{Cohn:2019:NAACL},
summarization \cite{Shen:2019:NAACL} and referring expression generation \cite{Zarriess:2019:ACL}.

However, its application to the dialogue domain remains understudied.
In this work, we explore how the RSA framework can be adopted in dialogue agents to alleviate the inconsistency problem.
Also, we further extend the framework by making the distractor selection as a learnable process. %

\section{{Insensitivity to Contradictory Words \newline in Existing Persona-based Agents}}
\label{sec:contradictions}

Although conditional language generation has shown promising progress,
maintaining consistency within the generation yet remains unsolved.
From quantitative evaluation, we reveal existing generative models for dialogues are highly insensitive to contradictory words.

\begin{figure}[t!] \begin{center}
    \includegraphics[width=\linewidth]{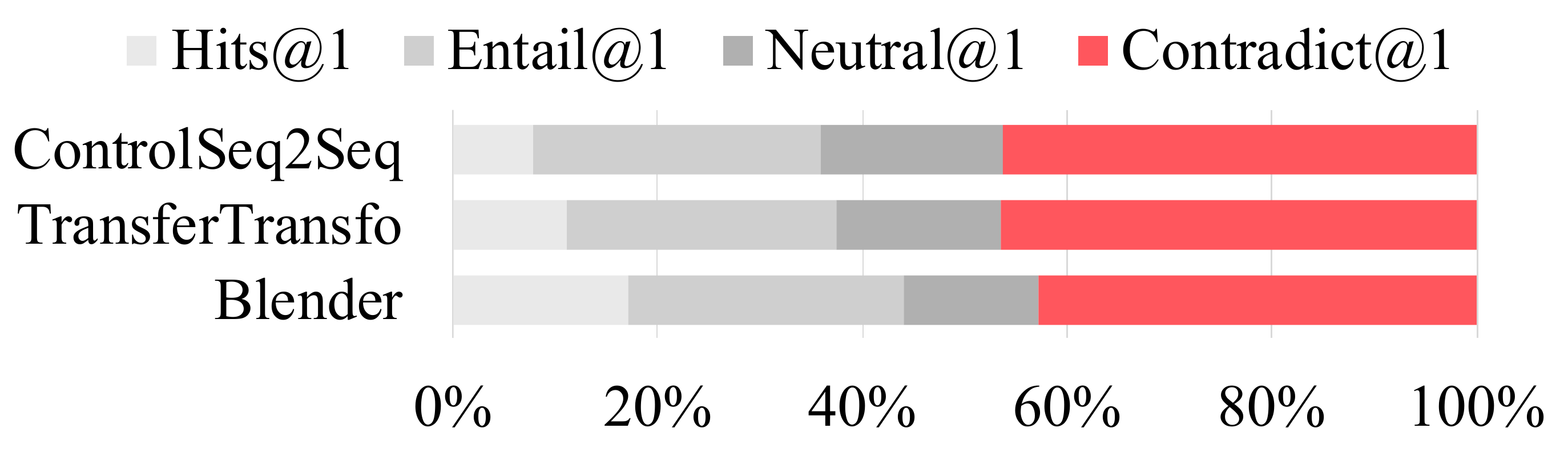}
    \vspace{-15pt}
    \caption{Proportion of Hits@1, Entail@1, Neutral@1 and Contradict@1 in the top-1 candidates returned by the models on the Dialogue NLI dataset.
    }
      \label{fig:contradiction_ratio}
    \vspace{-8pt}
\end{center} \end{figure}

{\renewcommand{\arraystretch}{1.1}%
    \begin{table}[t!] \begin{center}
    \small
    \setlength{\tabcolsep}{6pt}
    \begin{tabular}{lcccc}
        \toprule
                                 & ROUGE-1         & ROUGE-L          & SPICE \\
        \midrule
        \addlinespace[0.1cm]
        GT Utterance             & 15.7            & 14.6             & \textbf{10.6}   \\
        Top Entail-Utt           & 15.3            & 14.5             & 7.1    \\
        Contradict@1-Utt         & \textbf{16.3}   & \textbf{15.9}    & 6.6    \\
        \bottomrule
    \end{tabular}
    \vspace{-5pt}
    \caption{
        Comparison between ground-truth utterances, top-ranked entailing candidates and Contradict@1 utterances in ROUGE and SPICE scores. %
    }
    \vspace{-16pt}
    \label{tab:candidates_rouge}
\end{center}\end{table}}

\textbf{Dialogue NLI Evaluation}.
\citet{Welleck:2019:ACL} introduce the Dialogue NLI dataset based on the PersonaChat dataset \cite{Zhang:2018:ACL}. %
They collect entailing and contradictory utterances to the given persona, and release an evaluation set comprised of dialogues each with 31 utterance candidates:
10 entailing, 10 neutral, and 10 contradictory utterances with 1 ground-truth (GT) utterance.
On this evaluation set, we run three recent models \cite{See:2019:NAACL, Wolf:2019:arXiv, Roller:2020:blender} that achieve the best performance on PersonaChat.
We report four ranking metrics following \citet{Welleck:2019:ACL}: Hits@1, Entail@1, Neutral@1 and Contradict@1.
Each metric is the proportion of GT, entailing, neutral and contradictory utterances in the top-1 candidates returned by the model, respectively.
The models rank the candidates by perplexity scores.

Figure \ref{fig:contradiction_ratio} shows that all three models select contradictory candidates much more often than the GT utterances (see further results in Table \ref{tab:dnli_results}).
Though models are conditioned on a given persona, they are highly insensitive to contradictions.

{\renewcommand{\arraystretch}{1.2}%
    \begin{table}[t!] \begin{center}
    \small
    \setlength{\tabcolsep}{4pt}
    \begin{tabular}{cl}
        \toprule
        \addlinespace[0.2cm]
        Persona & \makecell[l]{I love wearing skinny jeans and shirts. \\ I am a blonde girl with short hair.} \\
        \addlinespace[0.1cm]
        \cmidrule(lr{0.3em}){1-2}
        \addlinespace[0.2cm]
        \makecell{GT Utterance} & \makecell[l]{(I, 1.87) (have, 51.42) (really, 201.45) \vspace{2pt} \\ (short, 1.78) (hair, 1.30) (and, 2.81) \vspace{2pt} \\(it, 45.25) (is, 2.19) (\textcolor{myblue}{blonde}, 461.60).}  \\
        \addlinespace[0.1cm]
        \cmidrule(lr{0.3em}){1-2}
        \addlinespace[0.2cm]
        \makecell{Contradict@1-Utt} & \makecell[l]{(What, 60.89) (color, 103.11) (is, 1.99) \vspace{2pt} \\ (your, 1.06) (hair, 1.05) (?, 1.11) \vspace{2pt} \\(Mine, 3.57) (is, 1.03) (\textcolor{myred}{brown}, 17.25).}  \\
        \addlinespace[0.1cm]
        \bottomrule
    \end{tabular}
    \caption{
        Example of a contradictory utterance returned by the model and its GT utterance with perplexity per token. %
        The words of entailment and contradiction to the persona are shown in blue and red, respectively.
    }
    \vspace{-17pt}
    \label{tab:contradict_example}
\end{center}\end{table}}

\subsection{Analysis of Contradict@1 Utterances}

To investigate why insensitivity to contradiction prevails in the state-of-the-art models, we further analyze the contradictory utterances returned by the models (Contradict@1-Utt), comparing with the GT utterances and the top-ranked entailing candidates (Top Entail-Utt). %
Table \ref{tab:candidates_rouge} reports language metrics between the selected candidates and the given persona sentences using SPICE \cite{Anderson:2016:spice} and ROUGE \cite{Lin:2004:rouge}.
SPICE metric measures semantic similarity and ROUGE metric measures $n$-gram overlaps between two sentences.
Contradict@1-Utt shows \textit{lower} SPICE scores and \textit{higher} ROUGE scores than other utterances, implying that it may be different in semantics but similar in syntax to the given persona.
To take a closer look, we extract the contradicting words from Contradict@1-Utt and their counterparts from GT utterances to compare their average perplexity scores.
In the Dialogue NLI dataset, every utterance is labeled with a triple $(entity_1, relation, entity_2)$,
such as ``\textit{I just like to listen to rock music}'' with $(i, like\_music, rock)$.
By construction, Contradict@1-Utt must contain words that are contradictory to the GT utterance and the given persona.
The perplexity scores of contradictory words (106.7) were considerably lower than those of the counterparts in GT utterances (280.1).
Table \ref{tab:contradict_example} shows an example of such dialogue instance with perplexity per word.
If properly conditioned with the given persona, models should show lower perplexity for the words in the persona.
However, their perplexity scores are significantly higher than those of contradictory words.
It reveals that models behave more as a plain language model rather than as a persona-conditioned model.
Thus, guarantee of consistency for each word generation step is required for persona-based dialogue agents to resolve such issue.

\section{Approach}
\label{sec:approach}

We introduce how to endow dialogue agents with public self-consciousness, which helps them keep consistency in mind at each generation step by reflecting an imaginary listener's distribution over personas.
Since the imaginary listener arises from the plain dialogue-agent, separate training is not needed.
Figure \ref{fig:model} illustrates its overall structure.

We present how to model public self-consciousness using the Rational Speech Acts (RSA) framework \cite{Frank:2012:Science} in Section \ref{sec:self_cons}.
We then discuss learning of distractor selection as our major novelty for the RSA in Section \ref{sec:learn_distractor}.

\subsection{Modeling the Public Self-Consciousness}
\label{sec:self_cons}
We seek to build a dialogue agent who is self-conscious about its consistency without the need for training on NLI labels or rating consistency with NLI models.
Given that modeling the interactions between listener and speaker is a main topic in pragmatics,
we take advantage of the RSA framework \cite{Frank:2012:Science}.
It treats language use as a recursive process where probabilistic speaker and listener reason about each other's intentions in a Bayesian fashion.
To apply the framework to sequence generation for dialogues, we extend the incremental approach proposed for image captioning \cite{Cohn:2018:NAACL}.

To generate an utterance, the agent computes the distribution of every next token $u_t$ at timestep $t$ in Bayesian fashion as follows. %

\begin{figure}[t] \begin{center}
    \vspace{-5pt}
    \includegraphics[width=\linewidth]{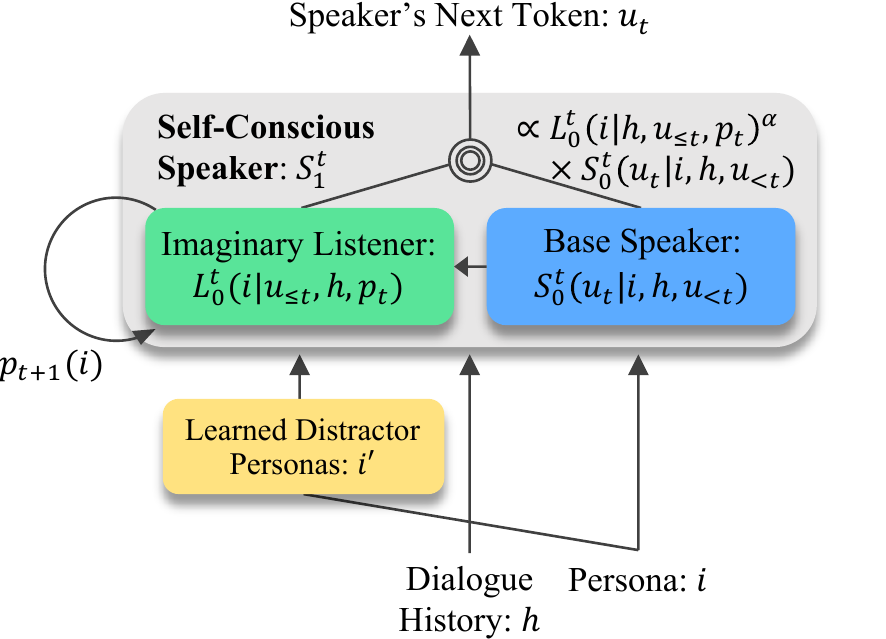}
    \caption{The proposed self-conscious agent $S_1$ consists of base speaker $S_0$ and imaginary listener $L_0$. It recursively generates the next token $u_t$ at every time $t$.}
    \label{fig:model}
    \vspace{-10pt}
\end{center} \end{figure}

\textbf{Base Speaker $S_0$.} We first assume persona $i$ is given to the base speaker, along with the dialogue history $h$ and partial utterance $u_{<t}$, as shown in Figure \ref{fig:model}.
The base speaker $S_0^t$ returns a distribution over the next token at timestep $t$: $S_0^t(u_t | i, h, u_{<t})$.
Any conditional dialogue agent can be used as a base speaker. See the details in Section \ref{sec:setting}.

\textbf{Imaginary Listener $L_0$.} While the base speaker generates each token one at a time, the imaginary listener reasons about the speaker's persona.
The imaginary listener $L_0^t$ is the posterior distribution of the speaker's persona  in terms of the base speaker and the world prior $p_t(i)$ over personas as follows,
\begin{multline} \label{eq:listener}
    L_0^t(i|h, u_{\leq t}, p_t) \\
    \propto \frac{S_0^t(u_t|i, h, u_{<t})^\beta \times p_t(i)}{\sum_{i' \in \mathcal{I}} S_0^t(u_t|i', h, u_{<t})^\beta \times p_t(i')}.
\end{multline}
\noindent %
where $\beta$ on $S_0^t$ is the listener rationality coefficient that controls the amount of information from the current timestep compared to the cumulative prior $p_t(i)$.
$L_0$ returns a probability distribution over the personas in world $\mathcal{I}$, which is a finite set ($|\mathcal{I}|=3$) comprising the given persona $i$ and distractor personas.
The distractors are different personas from other dialogue instances in the dataset.
We decide world $\mathcal{I}$ per dialogue instance through learning, which will be elaborated in Section \ref{sec:learn_distractor}.

\textbf{Self-Conscious Speaker $S_1$.} With $S_0^t$ and $L_0^t$, the self-conscious speaker $S_1^t$ is defined as
\begin{multline}
    S_1^t(u_t|i, h, u_{<t}) \\
    \propto L_0^t(i|h, u_{\leq t}, p_t)^\alpha \times S_0^t(u_t|i, h, u_{<t}),
    \label{eq:s1}
\end{multline}
\noindent where $\alpha$ is the speaker rationality coefficient that determines how much the likelihood is considered.
By taking the listener's distribution into account, the speaker is now self-conscious about what persona it sounds like.
Especially, the agent seeks to be perceived as the given persona $i$ rather than some other persona $i'$.
The likelihood of each token being identified as the persona $i$ acts as a bonus added to the base speaker's token scores.
Hence, tokens that are consistent to the given persona are preferred to others.
The token with the highest probability is added to the partial utterance, becoming the next input $u_{<t+1}$ for the speaker.

\textbf{Updating the world prior with $L_0$.} Starting from a uniform distribution as the initial prior $p_0(i)$, we  update the world prior $p_{t+1}(i)$ according to $S_1$'s output $u_t$ at every time step:
\begin{align}
    p_{t+1}(i) = L_0^t(i|h, u_{\leq t}, p_t).
    \label{eq:world_update}
\end{align}
\noindent Hence, $p_t(i)$ represents the cumulative state of the partial utterance up to $t$.
\citet{Cohn:2018:NAACL} report the prior update with $L_1 \propto S_0^t(u_t|i, h, u_{<t}) \times L_0^t(i|h, u_{\leq t}, p_t)$ makes little practical effect compared to a uniform prior.
We find that updating the prior with Eq. (\ref{eq:world_update}) instead is effective. See the results in Section \ref{sec:ablation}.

\subsection{Learning to Select Distractors}
\label{sec:learn_distractor}

Distractors \cite{Andreas:2016:EMNLP} are samples (\eg other personas in the dataset) which are different from the given target.
In previous works of RSA, the distractors to be included in world $\mathcal{I}$ are selected manually or randomly from the dataset.
However, we find that performance variance is large according to the selected distractors. %
We thus propose to learn distractor selection, especially based on the life-long memory network \cite{Kaiser:2017:ICLR}. 
The life-long memory network is capable of implicitly clustering similar dialogue contexts into a few slots with associated persona. %
Therefore, it can efficiently memorize and retrieve distractor personas for each context.
In Appendix, we experiment that our approach outperforms other models including BERT-based algorithms.

To better select useful distractor personas, supervised learning is desirable. %
However, there is no explicit label indicating which distractors are helpful for each dialogue.
We select the persona that have the best Hits@1 as the distractor label per training dialogue.
The Hits@1 is the score for favoring the ground-truth next utterance (consistent and context-relevant) over other candidate utterances which are just being consistent (\ie entailing) or contradictory to the given persona.
In other words, the score represents consistency and also appropriateness at the same time.
Thus, such distractors can help the self-conscious agent to generate responses which are context-relevant and allow the imaginary listener to identify the speaker's persona.
Each training datapoint comprises a given persona, a distractor persona and dialogue context.

\textbf{Memory Structure.}
The memory consists of three types of information: $M = (\mathbf K, \mathbf{v}, \mathbf{a})$. %
$\mathbf K \in \mathbb{R}^{m \times d}$ is a key matrix, where $m$ is the number of memory slots and $d$ is the dimension of the key vectors, which are the embedding of datapoints.
The value vector $\mathbf{v} \in \mathbb{R}^{m}$ stores the index of a persona. %
$\mathbf{a} \in \mathbb{R}^{m}$ is an age vector, which is used for memory update. %
We set $m = 16,000$ and $d = 768$.

\textbf{Memory Addressing.}
We construct the query vector $\mathbf q$ for each datapoint with the \textit{BERT-Uncased-Base} \cite{Devlin:2019:NAACL} model.
We use the output embedding of BERT's [CLS] token, and normalize it to a unit length to build $\mathbf q \in \mathbb R^d$. %

Using the cosine similarity between $\mathbf q$ and each memory key, we can find the $k$ nearest neighbors:
\begin{align}
    (n_1, n_2, ... , n_k) = NN_k(\mathbf{q}, \mathbf K).
\end{align}

\textbf{Memory Loss.}
Suppose that the query datapoint has a distractor label $l$.
Among $(n_1, ... , n_k)$, we denote the positive neighbor $n_p$ as the one with $\mathbf v[n_p] = l$
and the negative neighbor $n_b$ with $\mathbf v[n_b] \neq l$.
If there are multiple positive neighbors, we pick the one with the smallest memory index.
If no positive neighbor is found, we select a random key whose value is $l$.
For the negative neighbor, we select one randomly from $(n_1, ... , n_k)$.
We set $k=2048$.
Then, the loss is computed as
\begin{align}
    \mathcal{L} = \max(\mathbf q \cdot \mathbf K[n_b] - \mathbf q \cdot \mathbf K[n_p] + \alpha, 0),
\end{align}
where $\alpha$ is a positive margin, which we set as $0.2$.
This loss maximizes the cosine similarity between the query $\mathbf q$ and the positive key $\mathbf K[n_p]$, while minimizing the similarity to the negative key $\mathbf K[n_b]$.
We finetune the query network BERT with this loss.

\textbf{Memory Update.}
After computing the loss, memory $M$ is updated differently for two cases.
(1) If the top-1 neighbor's value (\ie persona) is correct ($\mathbf{v}[n_1] = l$), the key vector is updated as:
\begin{align}
    K[n_1] \leftarrow \frac{\mathbf{q} + K[n_1]}{\norm{\mathbf{q} + K[n_1]}}.
\end{align}
(2) Otherwise ($\mathbf{v}[n_1] \neq l$), we make a slot for the query; we find the oldest memory slot $n'$ according to the age vector $\mathbf a$ and write
\begin{align}
    K[n'] \leftarrow \mathbf{q}, \,\,\,\, \mathbf{v}[n'] \leftarrow l, \,\,\,\, \mathbf{a}[n'] \leftarrow 0.
\end{align}

\textbf{Training \& Inference.}
In our \textit{Distractor Memory} network, training corresponds to updating the memory and the parameters of the query network.

At inference, given a test example, we obtain the query by encoding the dialogue context and the persona using BERT.
We find $n$ nearest keys from the memory, and use their values (\ie persona indices) as the distractor personas.
We set $n=2$.

\section{Experiments}
\label{sec:exp}
\vspace{-5pt}

We show that our self-conscious framework can significantly improve consistency and accuracy of state-of-the-art persona-based agents on two benchmark datasets. %
We prove its effectiveness using both automatic and human evaluations.
We also show our framework can be generalized to improve consistency of dialogue context beyond persona.

\subsection{Datasets}
\label{sec:datasets}

\textbf{Dialogue NLI Evaluation Set} \cite{Welleck:2019:ACL}.
This dataset is based on PersonaChat with additional NLI annotations. %
Its main task is to rank next-utterance candidates given previous context.
For each dialogue, they collect 31 next-utterance candidates in respect to the given persona: 10 entailing, 10 neutral  and 10 contradicting candidates with 1 ground-truth utterance.
In total, the evaluation set includes 542 instances.

\textbf{PersonaChat dialogue} \cite{Zhang:2018:ACL}.
This dataset involves two interlocutors who are each given a persona and asked to get to know each other while playing their roles.
This task was the subject of the ConvAI2 competition \cite{Dinan:2019:arXiv} at NeurIPS 2018.
The competition version contains 17,878 chitchat conversations conditioned on 1,155 personas for training and 1,000 conversations conditioned on 100 personas for validation.

\subsection{Experimental Setting}
\label{sec:setting}

\textbf{Base Speakers.}
We experiment on three pretrained models including ControlSeq2Seq \cite{See:2019:NAACL}, TransferTransfo \cite{Wolf:2019:arXiv}, and Blender \cite{Roller:2020:blender} as base speakers ($S_0$) for our self-conscious agents ($S_1$).
The ControlSeq2Seq is a Seq2Seq model with attention trained on Twitter dataset \cite{Miller:2017:arXiv} and finetuned on PersonaChat.
TranferTransfo based on GPT \cite{Radford:2018:OpenAI} is the winner of the ConvAI2 competition in automatic evaluation.
Blender, a recently released generative dialogue model, is the state-of-the-art open-domain chatbot.
Our approach improves these base speakers by granting them the sense of self-consciousness.
We defer implementation details to Appendix.

\textbf{Evaluation Metrics.}
For Dialogue NLI, we report three ranking metrics introduced in the original paper: Hits@1, Entail@1, and Contradict@1.
Each metric is the proportion of GT, entailing, and contradictory utterances in the top-1 candidates returned by the model, respectively. 
High scores in Entail@1 and low scores in Contradict@1 indicate better consistency with the persona.

For PersonaChat, we report Hits@1, standard F1 score, perplexity and C score, following the ConvAI2 protocol.
Hits@1 is the accuracy of choosing the ground-truth next-utterance among 20 candidates as the models rank the candidates by perplexity.
The C score is a metric for dialogue consistency, introduced in \citet{Madotto:2019:ACL}.
It computes pairwise comparison between utterance $u$ and persona sentence $p_j$ with a pretrained NLI model.
The NLI model returns 1, 0, -1 for entailment, neutrality, and contradiction, respectively.
We sum the NLI scores across persona sentences per dialogue instance: $\mbox{C}(u) = \sum_j \mbox{NLI}(u, p_j)$.

{\renewcommand{\arraystretch}{1}%
\begin{table}[t!] \begin{center}
    \small
    \setlength{\tabcolsep}{6pt}
    \begin{tabular}{lccc}
        \toprule
        Model                                          & Hits@1 $\uparrow$   & Entail@1 $\uparrow$      & Contradict@1 $\downarrow$           \\
        \midrule
        \multicolumn{4}{l}{ControlSeq2Seq \cite{See:2019:NAACL}}                                                                 \\
        \addlinespace[0.1cm]
        \hspace{1mm}$S_0$                             & 7.9              & 27.9                    & 46.3                   \\
        \hspace{1mm}$S_1$                             & 10.5             & 36.4                     & 34.0                   \\
        \hspace{1mm}$S_1$+DM                         & \textbf{13.1}    & \textbf{40.8}               & \textbf{24.5}          \\
        \midrule
        \multicolumn{4}{l}{TransferTransfo \cite{Wolf:2019:arXiv}}                                                                      \\
        \addlinespace[0.1cm]
        \hspace{1mm}$S_0$                             & 11.1             & 26.4                        & 46.5                 \\
        \hspace{1mm}$S_1$                             & 17.5             & 40.4                        & 29.7                 \\
        \hspace{1mm}$S_1$+DM                         & \textbf{18.8}    & \textbf{45.8}               & \textbf{19.7}        \\
        \midrule
        \multicolumn{4}{l}{Blender \cite{Roller:2020:blender}}                                                                      \\
        \addlinespace[0.1cm]
        \hspace{1mm}$S_0$                             & 18.8             & 27.3                        & 42.4                 \\
        \hspace{1mm}$S_1$                             & 21.8            & 38.0                      & 30.6                 \\
        \hspace{1mm}$S_1$+DM                         & \textbf{22.5}    & \textbf{44.1}             & \textbf{19.6}        \\
        \bottomrule
    \end{tabular}
    \caption{
        Comparison of our approach $(S_1)$ with base speakers $(S_0)$ on the Dialogue NLI evaluation set \cite{Welleck:2019:ACL}.
        +DM is the \textit{Distractor Memory}. %
        High scores in Hits@1, Entail@1 and low scores in Contradict@1 imply better consistency.
    }
    \vspace{-16pt}
    \label{tab:dnli_results}
\end{center} \end{table}}

\subsection{Quantitative Results} \label{sec:dnli} %

\textbf{Results on Dialogue NLI.}
Table \ref{tab:dnli_results} compares the performance of dialogue agents on the Dialogue NLI evaluation set.
Our self-conscious agent $S_1$ significantly reduces Contradict@1 scores and increases the Entail@1 along with the Hits@1 accuracy of the literal agents $S_0$.
We remind that each entailing candidate shares the same annotated triple as the GT utterance.
In other words, they have similar semantics to the GT utterance and follow the given persona.
Thus, Entail@1 is a lenient version of Hits@1 \cite{Welleck:2019:ACL}.
The \textit{Distractor Memory} (DM) is better than random distractor selection for $S_1$ across all metrics.
It concludes that learned distractors are more effective than random distractors for pragmatic agents.

{\renewcommand{\arraystretch}{1}%
    \begin{table}[t] \begin{center}
    \small
    \setlength{\tabcolsep}{6.3pt}
    \begin{tabular}{lccccc}
        \toprule
        Model                       &            & Hits@1 $\uparrow$        & F1 $\uparrow$              & Perplexity $\downarrow$     & C $\uparrow$     \\
        \midrule
        \multicolumn{5}{l}{ControlSeq2Seq \cite{See:2019:NAACL}}                                                                              \\
        \addlinespace[0.1cm]
        \hspace{1mm}$S_0$           &            & 16.1                     & 17.0                       & \textbf{22.9}               & 0.45  \\
        \cmidrule(lr{1em}){1-2}
        \hspace{1mm}$S_1$           &            & 16.4                     & 16.9                       & 23.9                        & 0.54  \\
        \hspace{1mm}$S_1$+DM        &            & \textbf{16.7}            & \textbf{17.1}              & 23.9                        & \textbf{0.55}  \\
        \midrule
        \multicolumn{5}{l}{TransferTransfo \cite{Wolf:2019:arXiv}} \\
        \addlinespace[0.1cm]
        \hspace{1mm}$S_0$            &           & 16.2                     & 19.2                       & \textbf{17.6}               & 0.86  \\
        \cmidrule(lr{1em}){1-2}
        \hspace{1mm}$S_1$            &           & 17.5                     & 19.4                       & 19.1                        & 0.96  \\
        \hspace{1mm}$S_1$+DM         &           & \textbf{18.2}            & \textbf{19.5}              & 19.1                        & \textbf{0.97}  \\
        \midrule
        \multicolumn{5}{l}{Blender \cite{Roller:2020:blender}} \\
        \addlinespace[0.1cm]
        \hspace{1mm}$S_0$            &           & 27.6                     & 19.5                       & \textbf{12.0}               & 0.85  \\
        \cmidrule(lr{1em}){1-2}
        \hspace{1mm}$S_1$            &           & 28.8                     & 19.7                       & 13.2                        & 0.93   \\
        \hspace{1mm}$S_1$+DM         &           & \textbf{29.1}            & \textbf{19.8}              & 13.2                        & \textbf{0.95}  \\
        \bottomrule
    \end{tabular}
    \caption{
        Comparison of our approach $(S_1)$  with base speakers $(S_0)$ on PersonaChat \cite{Zhang:2018:ACL}.
        C is the consistency score evaluated by a pretrained NLI model \cite{Madotto:2019:ACL}.
        For TransferTransfo, we use the generative version to calculate Hits@1. %
    }
    \vspace{-8pt}
    \label{tab:personachat_results}
\end{center}\end{table}}

\textbf{Results on PersonaChat.} %
Table \ref{tab:personachat_results} compares the performance of different dialogue agents on the PersonaChat dataset.
Our model $S_1$ outperforms all other generative dialogue agents in terms of consistency related metrics, \ie Hits@1 and C score.
Since the posterior update of our self-conscious agent revises the distribution learned by the base speaker, the increase in perplexity is natural due to the effect of regularization.
Nevertheless, our approach improves the F1 score for TransferTransfo and Blender.
Thus, being consistent to the given persona can also help improve the generation performance of dialogue agents.

{\renewcommand{\arraystretch}{1}%
\begin{table}[t!] \begin{center}
    \small
    \setlength{\tabcolsep}{3pt}
    \begin{tabular}{lccc}
        \toprule
        Model                                          & Hits@1 $\uparrow$   & Entail@1 $\uparrow$      & Contradict@1 $\downarrow$           \\
        \midrule
        \multicolumn{4}{l}{ControlSeq2Seq \cite{See:2019:NAACL}}                                                                 \\
        \addlinespace[0.1cm]
        \hspace{1mm}$S_0$+NLI                       & 12.7             & 48.2                      & 8.1               \\ %
        \hspace{1mm}[$S_1$+DM]+NLI                    & \textbf{14.4}    & \textbf{51.7}             & \textbf{7.0}      \\ %
        \midrule
        \multicolumn{4}{l}{TransferTransfo \cite{Wolf:2019:arXiv}}                                                                      \\
        \addlinespace[0.1cm]
        \hspace{1mm}$S_0$+NLI                       & 17.2             & 44.4                      & 9.8               \\ %
        \hspace{1mm}[$S_1$+DM]+NLI                   & \textbf{21.4}     & \textbf{54.6}             & \textbf{5.4}      \\
        \midrule
        \multicolumn{4}{l}{Blender \cite{Roller:2020:blender}}                                                                      \\
        \addlinespace[0.1cm]
        \hspace{1mm}$S_0$+NLI                        & 24.9             & 44.7                      & 6.0        \\
        \hspace{1mm}[$S_1$+DM]+NLI                     & \textbf{26.6}    & \textbf{52.0}             & \textbf{5.7}        \\
        \bottomrule
    \end{tabular}
    \caption{
        Comparison of our approach $(S_1)$ with base speakers $(S_0)$ on the Dialogue NLI evaluation set \cite{Welleck:2019:ACL} with pretrained NLI model attached.
    }
    \vspace{-16pt}
    \label{tab:nli_labels_results}
\end{center} \end{table}}

\textbf{Comparison with agents that use NLI model.}
We also test agents with pretrained NLI models attached \cite{Welleck:2019:ACL}, denoted by +NLI in Table \ref{tab:nli_labels_results}.
The NLI model computes contradiction scores of each candidate utterances, and penalize its rank accordingly.
Compared to base agents with no self-consciousness, our agents improve consistency in all three metrics even further when using additional NLI models.
Another notable result is that our agents without NLI ($S_1$+DM in Table \ref{tab:dnli_results}) for ControlSeq2Seq and TransferTransfo even outperform the base agents with NLI ($S_0$+NLI) on Hits@1.
That is, our self-conscious agents achieve better GT accuracy even without the help of an NLI model trained on consistency labels.

{\renewcommand{\arraystretch}{1.2}%
    \begin{table}[t] \begin{center}
    \small
    \setlength{\tabcolsep}{3pt}
    \begin{tabular}{lcccc}
        \toprule
                                                    &  \multicolumn{2}{c}{Raw}                            & \multicolumn{2}{c}{Calibrated}            \\
         \cmidrule(lr{1em}){2-3} \cmidrule(lr{1em}){4-5}
        Model                                       & Consistent             & Engaging                   & Consistent      & Engaging           \\
        \midrule
        \multicolumn{5}{l}{TransferTransfo \cite{Wolf:2019:arXiv}}                                                           \\
        \addlinespace[0.1cm]
        \hspace{0.1mm}$S_0$                           & 0.53 (0.02)            &  2.48 (0.03)              & 0.44 (0.01)                &  2.48 (0.01)              \\
        \cmidrule(lr{0.02em}){1-1}
        \hspace{0.1mm}$S_1$+DM                        & \textbf{0.61} (0.02)   &  \textbf{2.55} (0.03)     & \textbf{0.52} (0.01)       & \textbf{2.52} (0.01)               \\
        \bottomrule
    \end{tabular}
    \caption{Human evaluation results comparing the consistency and engagingness of the base speaker ($S_0$) and our self-conscious agent ($S_1$). Numbers in parentheses are the standard errors.}
    \vspace{-7pt}
    \label{tab:human}
    \end{center}\end{table}}

\subsection{Human Evaluation} \label{sec:human}

We perform human evaluation via Amazon Mechanical Turk.
We random sample 250 test examples, each is rated by three unique human judges in terms of (i) \textit{Consistency} and (ii) \textit{Engagingness}.
Turkers are shown a given persona, a dialogue context, and the model's generated utterance.
For consistency, we follow \citet{Madotto:2019:ACL} and ask judges to assign $1$, $0$, $-1$ to the utterance for consistency, neutrality, and contradiction, respectively.
Following \citet{See:2019:NAACL}, we evaluate the engagingness of the utterance in a 4-point scale, where higher scores are better.
To alleviate annotator bias and inter-annotator variability, we apply Bayesian calibration \cite{Kulikov:2019:calibration} to the scores.

Table \ref{tab:human} summarizes the human evaluation results.
The agent with our self-consciousness method $S_1$ is rated as more consistent than the base agent $S_0$ while maintaining a similar level of engagingness.
While it can be trivial to increase consistency at the cost of engagingness (\eg perfect consistency can by generating boring utterances with very little variance),
it is not the case for our agent.
Since our agent seeks to be heard as the given persona to the listener, self-distinctive words tend to meld into generated responses (see Figure \ref{fig:alpha_example}).
Thus, the responses from self-conscious agents have their own color, which can help improving engagingness.

\begin{figure}[t!] \begin{center}
    \vspace{-2pt}
    \includegraphics[width=0.96\linewidth]{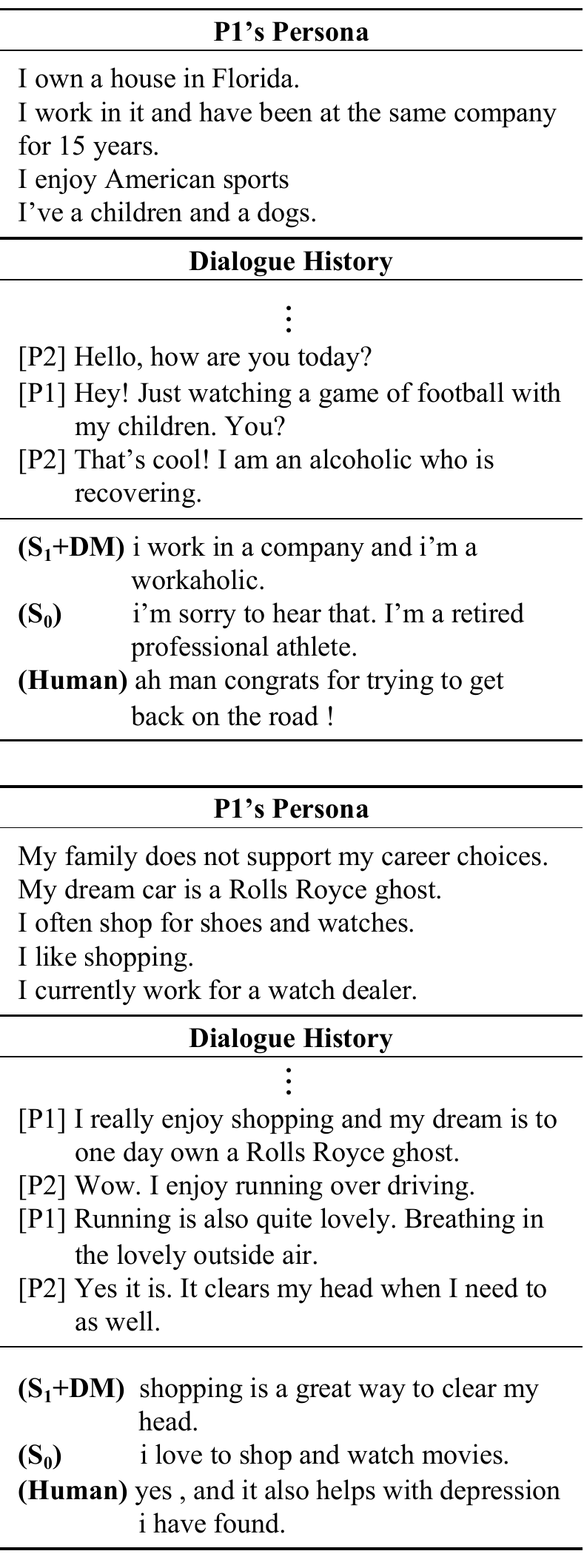}
    \caption{
        Examples of generated responses by our self-conscious agent with \textit{Distractor Memory} ($S_1$+DM) on the PersonaChat dataset \cite{Zhang:2018:ACL}.
        We compare it with the base speaker ($S_0$) of TransferTransfo \cite{Wolf:2019:arXiv} and the human response (Human).
    }
    \vspace{-5pt}
    \label{fig:generation_examples}
\end{center} \end{figure}

Figure \ref{fig:generation_examples} displays selected examples of utterance generation.
Each example is comprised of dialogue history, human response, and utterances generated by our method and baselines.

\subsection{Consistency for Dialogue Context} \label{sec:context}

We demonstrate that our self-conscious agent can be generalized to generate context-consistent utterances beyond persona.
We condition the agent with its previous responses in the dialogue history; that is, $i$ in Eq. (\ref{eq:s1}) is the agent's past responses instead of persona sentences.
Hence, tokens that are inconsistent to the agent's past response would be less favored by the model.

{\renewcommand{\arraystretch}{1}%
\begin{table}[t!] \begin{center}
    \small
    \setlength{\tabcolsep}{3.5pt}
    \begin{tabular}{lccc}
        \toprule
        Model                                       & Hits@1 $\uparrow$   & Entail@1 $\uparrow$        & Contradict@1 $\downarrow$            \\
        \midrule
        \multicolumn{4}{l}{Dialogue NLI \cite{Welleck:2019:ACL}}                                  \\ %
        \addlinespace[0.1cm]
        \hspace{0.5mm}$S_0$                          & 18.8              & 27.3                  & 42.4                 \\
        \hspace{0.5mm}$S_1$ (on context)             & \textbf{32.7}    & \textbf{27.7}         & \textbf{26.4}        \\
        \addlinespace[0.05cm]
        \bottomrule
    \end{tabular}
    \vspace{-8pt}
\end{center} \end{table}}

{\renewcommand{\arraystretch}{1}%
    \begin{table}[t!] \begin{center}
    \vspace{-9pt}
    \small
    \setlength{\tabcolsep}{5.1pt}
    \begin{tabular}{lcccc}
        \toprule
        \addlinespace[0.2cm]
        Model                                   & Hits@1 $\uparrow$        & F1 $\uparrow$              & Perplexity $\downarrow$     & C $\uparrow$     \\
        \midrule
        \multicolumn{5}{l}{PersonaChat \cite{Zhang:2018:ACL}}                                                                              \\
        \addlinespace[0.1cm]
        \hspace{0.5mm}$S_0$                       & 27.6                     & 19.5              & \textbf{12.0}               & 0.57  \\
        \hspace{0.5mm}$S_1$ (on context)          & \textbf{30.5}            & \textbf{19.9}     & 13.5                        & \textbf{0.58}  \\
        \midrule
        \multicolumn{5}{l}{EmpatheticDialogue \cite{Rashkin:2019:ACL}}                                                                            \\
        \addlinespace[0.1cm]
        \hspace{0.5mm}$S_0$                       & 32.6                     & 20.5                      & \textbf{14.7}               & 0.47  \\
        \hspace{0.5mm}$S_1$ (on context)          & \textbf{34.2}            & \textbf{20.6}              & 15.4                        & \textbf{0.50}  \\
        \bottomrule
    \end{tabular}
    \caption{
      Comparison of our approach $(S_1)$ with base speaker Blender $(S_0)$ when conditioned on dialogue context in three datasets.
      We compute the consistency score C respect to the dialogue context.
    }
    \vspace{-15pt}
    \label{tab:context_results}
\end{center}\end{table}}

Table \ref{tab:context_results} reports the results of context conditioned self-conscious agents.
The EmpatheticDialogue \cite{Rashkin:2019:ACL} is an open-domain dialogue dataset where a speaker describes a past emotional experience and the listener responds accordingly.
Since the speaker's descriptions should be consistent to the experience and previous utterances, it is a suitable benchmark for consistency.
We model the speaker's utterances and measure its consistency.

Our $S_1$ agent outperforms other literal agents on all three datasets in terms of consistency.
Thus, our approach can also be applied to help agents stay more consistent to its context.

\subsection{Controlling the Self-Conscious Agent} \label{sec:ablation}

To further analyze our self-conscious agent, we conduct experiments 
by controlling three features of our agent: world prior updates $p_t(i)$, listener rationality $\beta$ and speaker rationality $\alpha$.

\begin{figure}[t] \begin{center}
    \includegraphics[width=3.95cm]{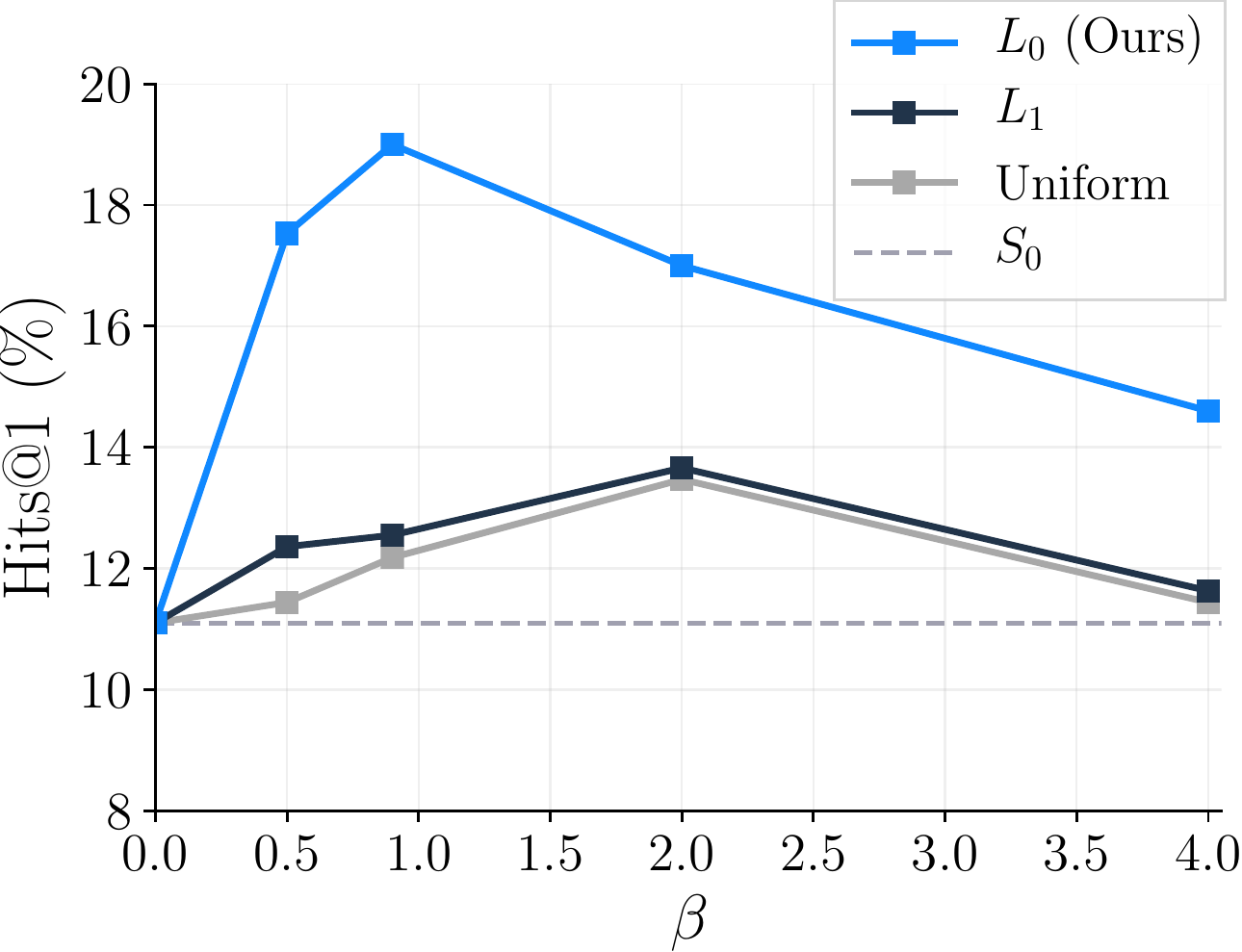}
    \includegraphics[width=3.65cm]{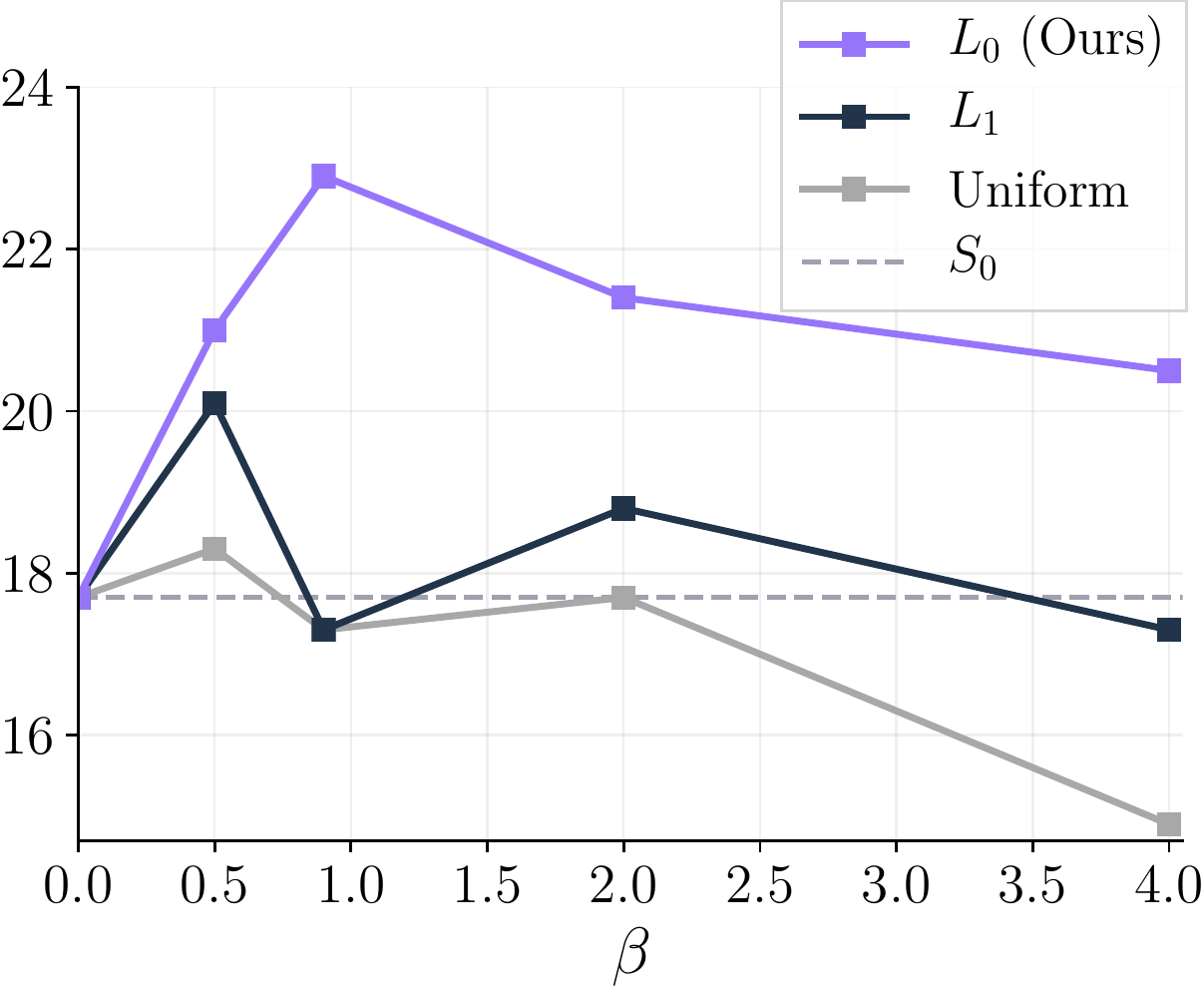}
    \caption{Performance variation of the self-conscious agent for TransferTransfo (left) and Blender (right) according to $\beta$.
      We compare different methods of updating the world prior $p_t(i)$ with $L_0$ (Ours), $L_1$ and a uniform prior.
             The dashed line is the base speaker $S_0$.}
    \label{fig:beta}
    \vspace{-10pt}
\end{center} \end{figure}

\textbf{World Prior Update.}
In the self-conscious agent, the world prior acts as a cumulative state over personas.
We remind that we propose to update the world prior with $L_0^t$ instead of $L_1^t$ in Eq. (\ref{eq:world_update}).
As reported in \citet{Cohn:2018:NAACL}, our experiments on the Dialogue NLI dataset confirm the prior update with $L_1^t$ makes little difference in performance compared with using a uniform distribution.
However, our approach with $L_0^t$ makes significant difference, as shown in Figure \ref{fig:beta}.
The reason is that the pragmatic listener $L_1^t \propto S_0^t(u_t|i, h, u_{<t}) \times L_0^t(i|h, u_{\leq t}, p_t)$ reflects the \textit{current} $S_0^t$ twice (\ie in $L_0^t$ and in itself) per time step.
Hence, the update with $L_1^t$ becomes more of an instantaneous prior rather than a cumulative one. %
On the other hand, $L_0^t$ moderately combines the information from both $S_0^t$ and $p_t(i)$, preserving better cumulative information.

\textbf{Listener Rationality $\beta$.}
We add $\beta$ in $L_0^t$ to control the amount of information incorporated to the world prior $p_t(i)$.
Figure \ref{fig:beta} depicts that when $\beta$ is large, the Hits@1 scores (\ie the GT accuracy) drop.
With a big $\beta$, the information $S_0^t$ at current time step overrides the cumulative prior $p_t(i)$.
That is, the utterance state evolves shortsightedly, ignoring the context information from the previous steps.
Therefore, setting of $\beta \le 1$ is advantageous for the self-conscious agent to incrementally decode.

\begin{figure}[t] \begin{center}
    \includegraphics[width=\linewidth]{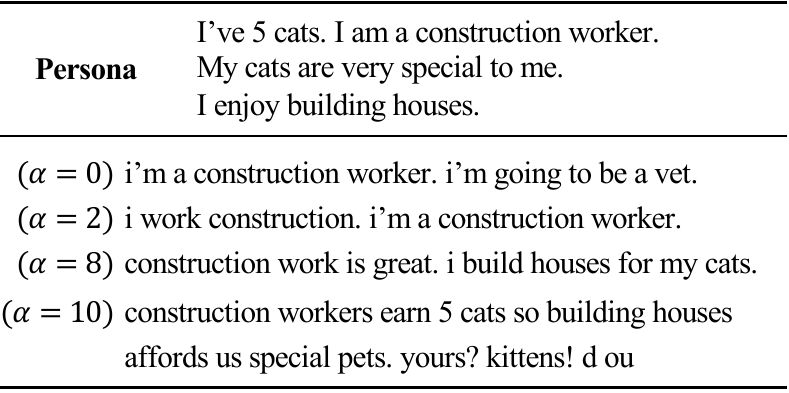}
    \vspace{-15pt}
    \caption{An example of utterance changes by controlling the speaker rationality $\alpha$ on the PersonaChat.}
    \label{fig:alpha_example}
    \vspace{-16pt}
\end{center} \end{figure}

\textbf{Speaker Rationality $\alpha$.}
Figure \ref{fig:alpha_example} shows an example of how generated responses vary according to the intensity of speaker rationality $\alpha$.
As $\alpha$ increases, the self-conscious agent reflects the listener's distribution (\ie the likelihood) more into the posterior.
When $\alpha$ is too large, the posterior distribution is overwhelmed by the likelihood of the persona.
Then, the language model degenerates to favor uttering fragments of the given persona while even ignoring the syntax.
Hence, $\alpha$ can control the degree of copying the given condition text.
An appropriate $\alpha$ value allows the given persona condition to blend smoothly in the utterance.

\section{Conclusion}
\vspace{-5pt}
This work investigated how modeling public self-consciousness can help dialogue agents improve persona-consistency.
We showed existing dialogue agents are highly insensitive to contradiction, and introduced an orthogonally applicable method using the RSA framework \cite{Frank:2012:Science} to alleviate the issue.
We also designed a learning method for distractor selection, named \textit{Distractor Memory} and proposed a better update for the listener's world prior.
Furthermore, we demonstrated how our approach can be generalized to improve dialogue context-consistency.
Our self-conscious agents improved the base agents on the Dialogue NLI \cite{Welleck:2019:ACL} and PersonaChat \cite{Zhang:2018:ACL} dataset,
without consistency labels and NLI models.
An important future direction will be generating the distractors and learning the rationality coefficients.

\subsubsection*{Acknowledgements}

We would like to thank Reuben Cohn-Gordon, Sean Welleck, Junhyug Noh and Jiwan Chung for their valuable comments.
We also thank the anonymous reviewers for their thoughtful suggestions on this work.
This research was supported by Brain Research Program by National Research Foundation of Korea (NRF) (2017M3C7A1047860), Institute of Information \& communications Technology Planning \& Evaluation (IITP) grant funded by the
Korea government (MSIT) (No. 2017-0-01772, Video Turing Test, No. 2019-0-01082, SW StarLab), and Creative Pioneering Researchers Program through Seoul National University. Gunhee Kim is the corresponding author.

\bibliography{emnlp2020_selfconscious}
\bibliographystyle{acl_natbib}

\vfill\eject
\appendix

\section{Results on Variants of \newline Distractor Selection (Section \ref*{sec:learn_distractor})}
\label{sec:distractor_models}

{\renewcommand{\arraystretch}{1}%
    \begin{table}[h] \begin{center}
    \small
    \setlength{\tabcolsep}{2.7pt}
    \begin{tabular}{lccc}
        \toprule
        Model                                    & Hits@1 $\uparrow$  & Entail@1 $\uparrow$    & Contradict@1 $\downarrow$      \\
        \midrule
        \multicolumn{4}{l}{ControlSeq2Seq \cite{See:2019:NAACL}}                                               \\
        \addlinespace[0.1cm]
        \hspace{1mm}Random                       & 8.5                & 32.8                   & 37.6                  \\
        \hspace{1mm}Nearest                      & 7.6                & 32.8                   & 36.5                  \\
        \hspace{1mm}Farthest                     & 9.4                & 33.6                   & 35.4                  \\
        \hspace{1mm}BERT-Classifier              & 9.2                & 33.6                   & 35.6                  \\
        \hspace{1mm}BERT-Ranker                  & 9.6                & 33.3                   & 35.1                  \\
        \hspace{1mm}DM                           & \textbf{11.1}      & \textbf{36.0}          & \textbf{28.2}         \\
        \bottomrule
    \end{tabular}
    \caption{
        Quantitative results of the proposed \textit{Distractor Memory} (DM) and other distractor selection methods on the Dialogue NLI evaluation set \cite{Welleck:2019:ACL}.
    }
    \vspace{-10pt}
    \label{tab:distractor_results}
\end{center}\end{table}}

We compare our proposed \textit{Distractor Memory} (DM) with three heuristic methods, and two variants of the pretrained BERT model \cite{Devlin:2019:NAACL}.
As a straightforward baseline, we randomly select $k$ personas from training set and directly use it as distractors.
Second, we test the $k$-nearest search by speaker's persona, denoted by Nearest; for a given persona descriptions, we find its closest training persona embedding using cosine similarity on average pooled BERT features. 
The third baseline denoted by Farthest is to find the $k$-farthest persona among the training personas. 

We also compare with two variants of the BERT model.
The first variant is BERT-Classifier, which  takes dialogue context as input and returns the index of persona from training set as output.
The second variant is bi-encoder ranking model of \citet{Miller:2017:arXiv},  denoted by BERT-Ranker.
It encodes dialogue context and candidate persona with separate BERT encoders measuring its ranking with cosine similarity.
For both methods, we use top-$k$ ranked personas as distractors and set $k=4$ for all the methods.
We use Adam optimizer \cite{Kingma:2015:ICLR} with learning rate 2e-5 and finetune \textit{BERT-Uncased-Base} up to 3 epochs.

Table \ref{tab:distractor_results} compares the performance of different distractor selecting methods on the Dialogue NLI evaluation set \cite{Welleck:2019:ACL}.
We set $\alpha=8$, $\beta=0.5$, and $|\mathcal{I}|=5$.
The DM model outperforms all the baselines across all metrics.
The Farthest shows better performance than the Nearest.%
It can be understood that dissimilar distractors are more effective in the Rational Speech Acts framework \cite{Frank:2012:Science}.
The BERT-Ranker performs the best among baselines, but not as good as ours, which validates that memorization capability is effective for selecting useful distractors.

\section{Implementation Details}
\label{sec:implementation}

\textbf{Base Codes and Datasets.}
We use the ParlAI framework\footnote{\url{https://parl.ai/}} \cite{Miller:2017:arXiv} and HuggingFace's Transformers\footnote{\url{https://huggingface.co/transformers/}} \cite{Wolf:2019:arXivHugging} to implement our models and baselines.
We use Dialogue NLI \cite{Welleck:2019:ACL} and PersonaChat \cite{Zhang:2018:ACL} datasets from the ParlAI framework as is.
We use the default preprocessing in ParlAI.

\textbf{Training.}
Our self-consciousness approach improves consistency for any pretrained dialogue-agents without additional consistency labels and pretrained NLI models.
Since it post-processes the output probability of pretrained dialogue-agents in a Bayesian fashion, no additional model parameters are added to the dialogue agents.
Thus, it does not require any training.
In the case of using the Distractor Memory (DM), first we initialize \textit{BERT-Uncased-Base} with pretrained weights and finetune it up to 3 epochs with Adam optimizer with learning rate 2e-5.
Then we find the best distractor persona for each model and use those labels to train our DM.
We train our DM on one NVIDIA TITAN Xp GPU up to 7 epochs.

\textbf{Hyperparameters.}
For Dialogue NLI evaluation, we set the speaker rationality $\alpha = 8.0$, the listener rationality $\beta = 1.0$, and the cardinality of the world $\mathcal{I}$ to 3.
In PersonaChat evaluation, we set $\alpha = 2.0$, $\beta = 0.3$ for ControlSeq2Seq \cite{See:2019:NAACL},
$\alpha=2$, $\beta=0.9$ for TransferTransfo \cite{Wolf:2019:arXiv}, and $\alpha=2.0$, $\beta=0.5$ for Blender 90M \cite{Roller:2020:blender}.
We also set $|\mathcal{I}| = 3$.
We experiment $\alpha=\{1.0, 2.0, 4.0, 8.0, 16.0 \}$, $\beta=\{0.3, 0.5, 0.9, 1.0, 2.0, 4.0 \}$, and $|\mathcal{I}|=\{2, 3, 5\}$.
We choose the hyper-parameter configuration showing the best performance in Hits@1 for Dialogue NLI and F1 score for PersonaChat.
The posterior distribution of our self-conscious agents are computed deterministically.
For our Distractor Memory, we set the memory key matrix as $\mathbf K \in \mathbb{R}^{m \times d}$, where $m=16000$ and $d=768$.
We set the number of nearest neighbor $k=2048$.

\textbf{Inference.}
We use greedy decoding for all methods.
The average runtime for our self-conscious approach is dependent on the base dialogue agents and the cardinality of world $\mathcal{I}$ which can be run in parallel like beam search.

\textbf{Evaluation.}
We follow the evaluation of the ParlAI framework.
Following \citet{Madotto:2019:ACL}, we use the finetuned BERT-based NLI model\footnote{\url{https://github.com/HLTCHKUST/PAML}.} to compute the C score.

\section{Dialogue Examples}
\label{sec:examples}

\begin{figure*}[t!] \begin{center}
    \includegraphics[width=0.9\linewidth]{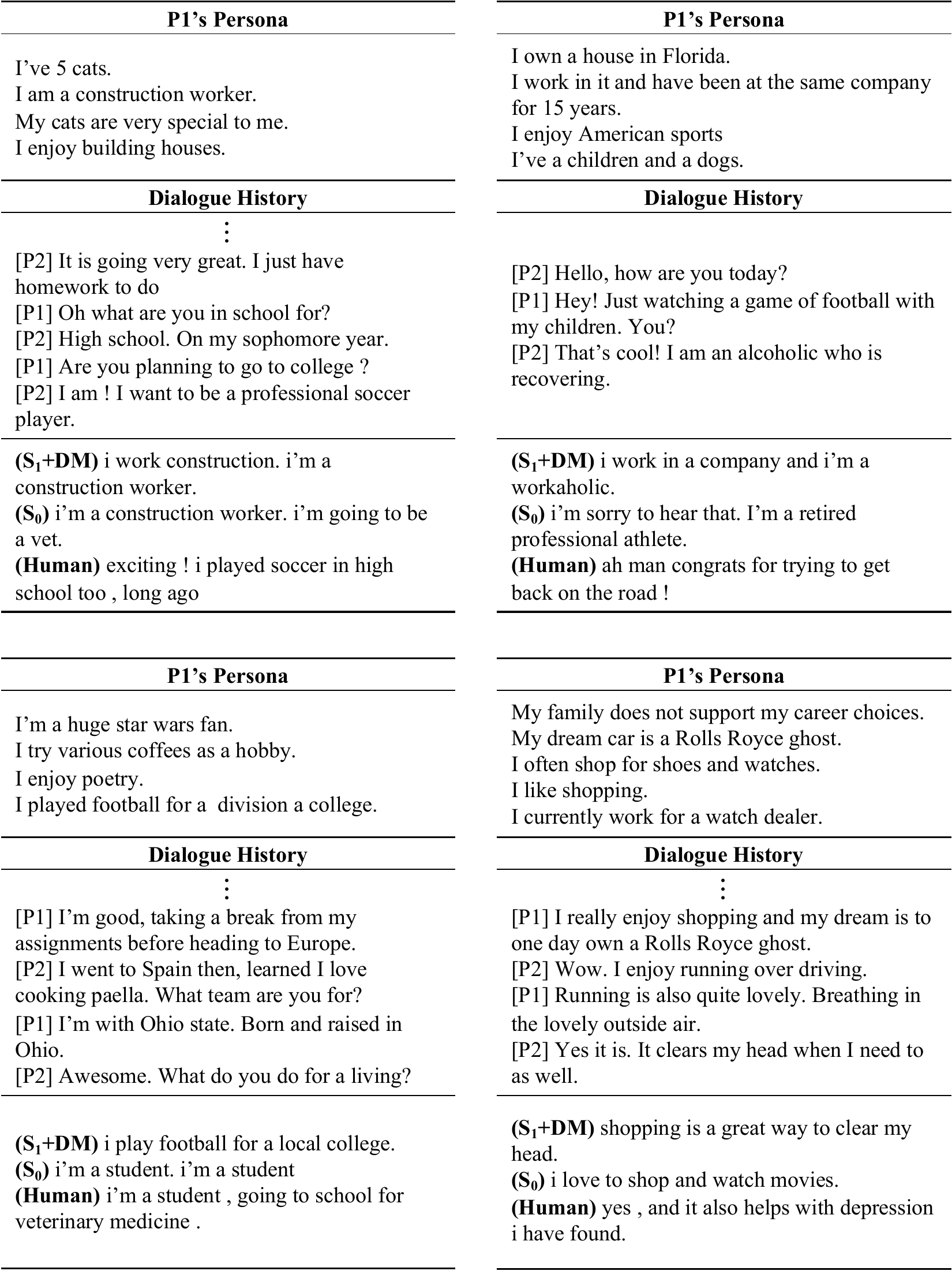}
    \caption{
        Examples of generated responses by our self-conscious agent with \textit{Distractor Memory} ($S_1$+DM) on the PersonaChat dataset \cite{Zhang:2018:ACL}.
        We compare it with the base speaker ($S_0$) of TransferTransfo \cite{Wolf:2019:arXiv} and the human response (Human).
    }
    \label{fig:figure_examples}
\end{center} \end{figure*}

Figure \ref{fig:figure_examples} shows selected examples of generated responses.
In each set, we show given persona, dialogue context, human responses, and generated responses by our self-conscious agent and the base speaker.
We use TransferTransfo \cite{Wolf:2019:arXiv} as a base speaker.

\end{document}